\def\year{2021}\relax
\documentclass[letterpaper]{article} 
\usepackage{aaai21}  
\usepackage{times}  
\usepackage{helvet} 
\usepackage{courier}  
\usepackage[hyphens]{url}  
\usepackage{graphicx} 
\usepackage{natbib}  
\usepackage{caption} 
\usepackage{amsmath}
\usepackage{amsfonts}
\usepackage{caption}
\usepackage{eucal}

\title{Anytime Prediction as a Model of Human Reaction Time }

\author{
    Omkar Kumbhar, \textsuperscript{\rm 1}
    Elena Sizikova, \textsuperscript{\rm 2}
    Najib Majaj, \textsuperscript{\rm 3}
    Denis G. Pelli, \textsuperscript{\rm 3}.
    \\
}
\affiliations{

    \textsuperscript{\rm 1}New York University, New York, United States\\
                           
    \textsuperscript{\rm 2}Center for Data Science, New York University, New York, United States\\
                           
    \textsuperscript{\rm 3}Department of Psychology, New York University, New York, United States\\
                           
    omkar.kumbhar@nyu.edu, es5223@nyu.edu, najib.majaj@nyu.edu, denis.pelli@nyu.edu
}

\nocopyright

\begin{document}

\maketitle

\begin{abstract}
Neural networks today often recognize objects as well as people do, and thus might serve as models of the human recognition process. However, most such networks provide their answer after a fixed computational effort, whereas human reaction time varies, e.g. from 0.2 to 10 s, depending on the properties of stimulus and task. To model the effect of difficulty on human reaction time, we considered a classification network that uses early-exit classifiers to make anytime predictions. Comparing human and MSDNet accuracy in classifying CIFAR-10 images in added Gaussian noise, we find that the network equivalent input noise SD is 15 times higher than human, and that human efficiency is only 0.6\% that of the network. When appropriate amounts of noise are present to bring the two observers (human and network) into the same accuracy range, they show very similar dependence on duration or FLOPS, i.e. very similar speed-accuracy tradeoff. We conclude that Anytime classification (i.e. early exits) is a promising model for human reaction time in recognition tasks.
\end{abstract}

\noindent \section{Introduction}

This project models the reaction time of human object recognition. There have been great advances in understanding and modeling how people recognize, but less on the timing. Dyslexia is a timing deficit. Dyslexia is defined as reading much more slowly than peers matched for age and education. It affects about 25\% of the population. Dyslexia impairs work and education. Current interventions, primarily phonemic awareness training, help significantly, but the problem remains. Further progress is stymied by lack of understanding of the nature of the underlying deficit that results in slow reading. Reading consists of successive eye fixations, each fixation allowing recognition of a word or partial word. Our project focuses on modeling and understanding the reaction time of that recognition. An accurate computational model of dyslexia represents a hypothesis about how the dyslexic brain processes text. 
Slow reading in dyslexic children has been previously attributed to a general deficit in stimulus classification speed and a linguistic deficit in lexical access speed\cite{pmid8177961}. 
While analysing eye tracking data between typical and dyslexia readers, it was seen that individuals with dyslexia would fixate for a longer time on each word as compared to typical readers\cite{DBLP:journals/corr/abs-1903-06274}. This is a crucial motivation for having a valid computational model of mimicking the human ability to detect objects in variable time and accuracy. 

The human ability to effortlessly recognize objects has long been a challenge to science and engineering.  Inspired by human neurophysiology, today's neural networks are often better than people at recognizing objects ~\cite{lecun2015deep,majaj2018deep}. These networks seem promising as the beginning of an explanation of how humans recognize, and, ultimately, how the human brain works. However, when there is little time, humans can trade off accuracy to respond more quickly, and this ability remains unexplained.  Here, we study anytime prediction ~\cite{horvitz2013reasoning}, i.e. the ability to respond after fewer FLOPs (floating point operations) with reduced accuracy, to model human reaction time.

As signal strength (e.g. contrast) increases, humans respond more quickly and more accurately, and there is a tight relation between sensitivity measured by accuracy or by reaction time\cite{palmer2005effect}. They showed that a diffusion model of perceptual decision making could account for the relation. Humans respond to instructions that change the emphasis on speed vs. accuracy, and can even learn to always respond with a fixed latency~\cite{pmid10641310}. We adopt that paradigm here. On each block of trials the observer is taught to respond at a fixed latency, which changes from block to block. Each block yields a point in a plot of accuracy vs. latency, and the many blocks trace out the speed-accuracy trade off. We measure network and human accuracy for the same stimuli and tasks, and measure the reaction time in milliseconds (ms) for the human, and calculate the number of floating point operations (FLOPs) consumed by the network. 
We analyse a recent classification model~\cite{huang2017multiscale} (MSDNet) that implements anytime prediction via early exits. MSDNet is designed to use less computation for easy images, to reduce the average computational load. MSDNet has intermediate classifiers with dense connectivity and multi-scale features as its main architectural design. Early exits are performed at these intermediate classifiers based on the confidence score of the output. 

The task is to identify the category (1 of 10) of an image from the CIFAR-10 set~\cite{Krizhevsky09learningmultiple}. To get human and machine accuracies that span a wide range, we test with various amounts of white Gaussian noise added to each image. It is known that the mathematically ideal classifier tolerates roughly 3 times higher noise SD as the human when identifying letters in common fonts~\cite{pelli2006feature}. Thus, we measure human and network reaction time for object recognition with added Gaussian noise, and compare their speed-accuracy trade-off curves. Our results indicate that anytime prediction is a promising model for accuracy and reaction time of human object recognition.

\section{Related Work}
\paragraph{Measuring the speed-accuracy tradeoff.} 
~\citeauthor{pmid10641310} analyzed the speed-accuracy trade-off in humans on a visual search task, where observers tried to find a target in an array of distractors. Task difficulty was increased by increasing set size. 

\begin{figure}[h!]
\begin{center}
\centerline{\includegraphics[width=\columnwidth]{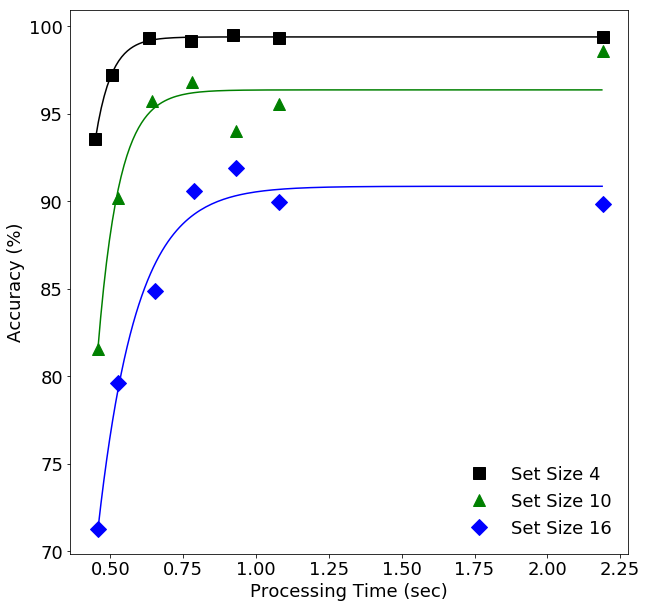}}
\caption{{\fontfamily{phv}\fontsize{9}{10}\selectfont
Human accuracy vs processing time~\cite{pmid10641310}. Human observers classified target orientation as same as or different from several distractors. Set size varied from 4 to 16 items, which increased the task difficulty. 
We fit a curve to the measured proportion correct $P$, defined by: 
  \begin{minipage}{\linewidth}
\begin{equation}
 P = 
  \begin{cases} 
    \lambda(1 - e^{-\beta(t-\delta)}) & \text{if } t>\delta \\\nonumber
   0       & \text{otherwise.} \nonumber
  \end{cases}
\end{equation}\nonumber
  \end{minipage}
\noindent This equation specifies accuracy as a function of processing time. $\lambda$ is asymptotic accuracy, $\delta$ and $\beta$ are the intercept and accuracy rate parameters. 
The fit parameters for set size 4 are: $\lambda=0.99, \delta=0.29, \beta=18$; Set size 10: $\lambda=0.96, \delta=0.32, \beta=14$; Set size 16: $\lambda=0.90, \delta=0.25, \beta=8. $}} 
\label{fig:analysis2}
\end{center}
\end{figure}

Figure~\ref{fig:analysis2} shows human accuracy as a function of processing time.

\paragraph{Anytime prediction in machine learning.}


The anytime prediction property ~\cite{pmlr-v22-grubb12} allows a network to classify an example $x$ within a finite computational budget $B>0$ (typically expressed in FLOPs). 
The Adaptive Computation Time (ACT)~\cite{DBLP:journals/corr/Graves16} algorithm allows recurrent neural networks to dynamically adapt to the needs of data, and learns the computational needs required to process an input. The authors of ACT were able to probe the structure of data and understand where the need of computation was more in transitions which were difficult to infer. 
~\citeauthor{Karayev_2014_CVPR} used a Markov Decision Process to model feature selection policy under an Anytime objective. Their work was mainly focused towards practical applications of vision, but they expected this model to be analysed for understanding human cognition. Anytime prediction is routinely used for time-sensitive applications such as pedestrian detection in self driving cars~\cite{cho2012real-time} and where accuracy is traded off for lower processing time, since very high accuracy at the expense of too many FLOPs is of no use if the car fails to detect a pedestrian in time. 

\paragraph{Modelling human reaction times using computational models.}
~\citeauthor{pmid23419619} propose a model to predict reaction time in response to natural images. This model is based on statistical properties of natural images and claimed to accurately predict human reaction time by defining a feature vector of entropy, Weibull~\cite{1951JAM....18..293W} $\beta$ and $\gamma$ parameters for each image and mapped it to the subject's reaction time using a linear equation.  ~\citeauthor{pmid14756592} use a speed-accuracy trade-off, response time and drift rate, defined as a function of the quality of information after processing the stimulus, to explain a lexical decision task (i.e. how rapidly does a person classify stimuli as words or non-words). Reaction time has been extensively studied in the context of perceptual decision making~\cite{palmer2005effect,wong2006recurrent,wagenmakers2007ez,basten2010brain}.

\section{Overview}
We train the MSDNet network on images to which noise might be added with probability 0.5. The best trained model is evaluated on test images with varying levels of Gaussian noise. After evaluating top-1 test accuracy of the test images, we analyse effects of noise on accuracy of early exit classifiers in the network. 
We also implement an experiment to record human reaction time on the task of object recognition on CIFAR-10 images. Additionally, we test one of the observers on varying noise of images. We further compare humans and machines based on the addition of noise and strike a correspondence between MFLOPs and time.


\paragraph{Images.}
We use the CIFAR-10 images ~\cite{Krizhevsky09learningmultiple} for all our experiments, with the default train/test split. This image set contains 50,000 training images and 10,000 test images of 32x32 pixels, and has 10 classes: airplane, automobile ,bird, cat, deer, dog, frog, horse, ship and truck.

\begin{figure}[t]
      \centering
      \vspace{0 ex}
      \includegraphics[width=\columnwidth]{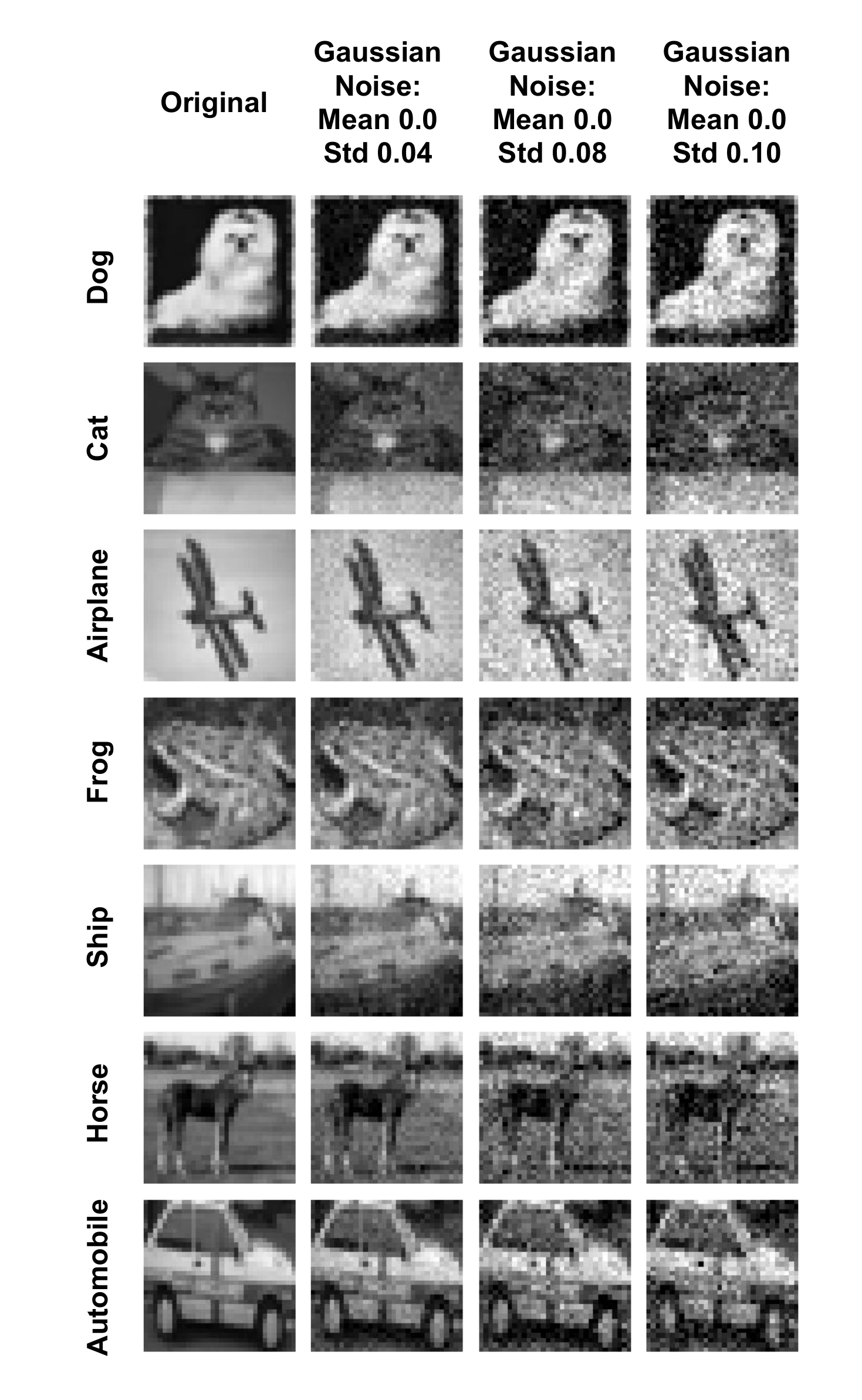}
      \vspace{-0 ex}
      \caption{{\fontfamily{phv}\fontsize{9}{10}\selectfont
      Effects of grayscale gaussian noise on sample images from CIFAR-10 dataset.}}
      \label{fig:noise}
      \vspace{-2 ex}
\end{figure}

\begin{figure}[t]
      \centering
      \includegraphics[width=\linewidth]{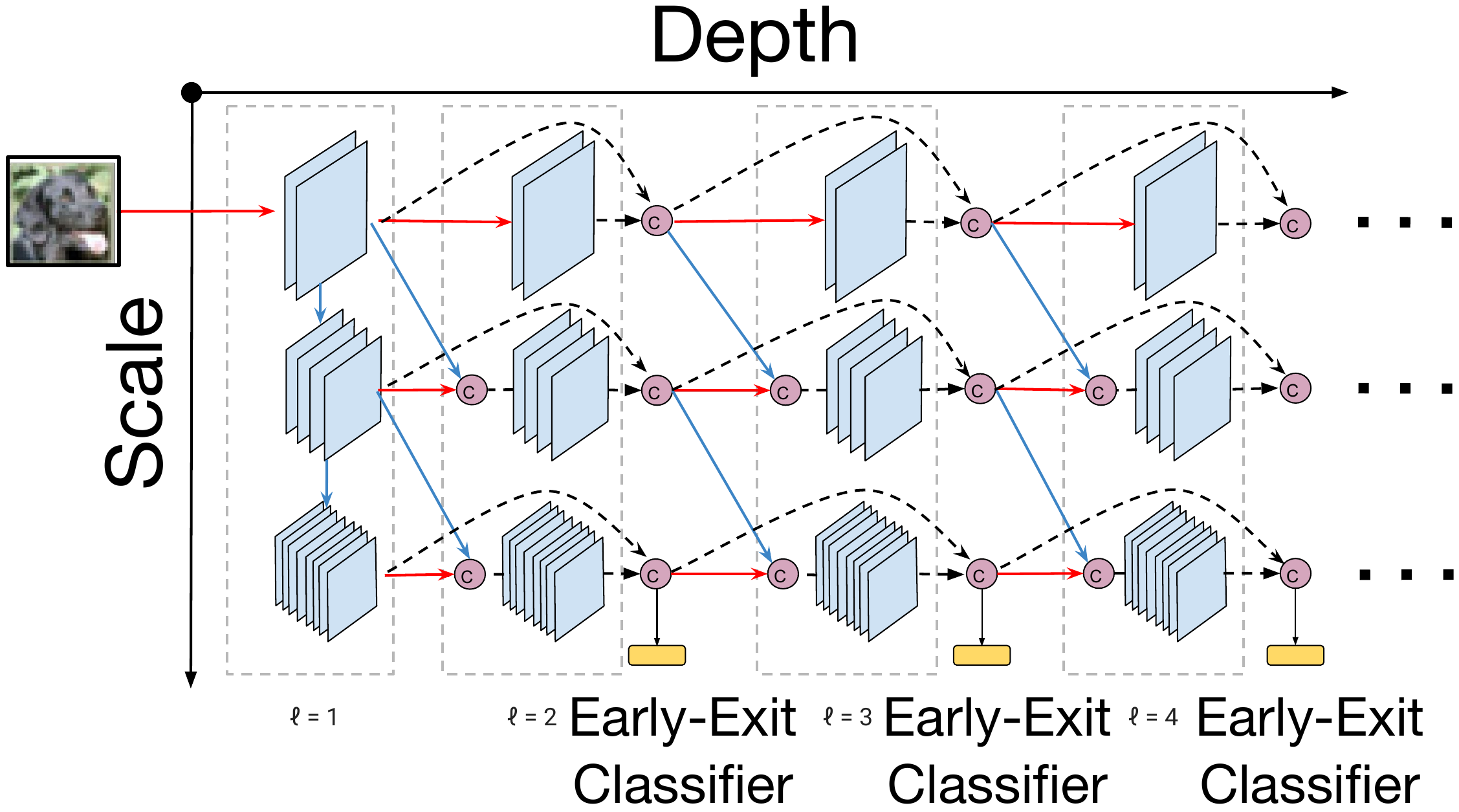}
      \vspace{-0 ex}
      \caption{{\fontfamily{phv}\fontsize{9}{10}\selectfont
      Illustration of MSDNet with three scales and four layers. Horizontal direction is the depth of the network and the vertical direction shows the scales. Horizontal arrows indicate a regular convolution operation, whereas diagonal and vertical arrows indicate a strided convolution operation. Early exits are shown in the form of intermediate classifiers which output the prediction when FLOPs are exhausted.}
      }
      \label{fig:layout}
\end{figure}

\paragraph{Human performance.}
Each image in the CIFAR-10 image set was labelled by human observers to make sure that the labelled picture is a prominent instance of a class, and clearly answerable. ~\cite{Krizhevsky09learningmultiple} Each image was converted into grayscale and then we added zero-mean white Gaussian noise with specified standard deviation. We used lab.js ~\cite{henninger_felix_2020_3953072} and JATOS ~\cite{Lange_2015} to create and deploy a survey to test humans on CIFAR-10.

\section{Method}
We followed a similar experimental protocol as ~\cite{pmid10641310} to design and conduct our object recognition experiment in humans. 
\paragraph{Observers.} Five observers, whose ages ranged from 20 to 25 years old, agreed to participate in an hour long session. Except for one observer (Human\#2), all observers were unsuspecting to the purpose and method of the experiment. Each observer had a normal or corrected-to-normal vision.
\paragraph{Stimuli.} The stimuli were presented via JATOS survey via worker links to each observers. Each observer had to perform a session of object detection on low noise (std 0.04) CIFAR-10 stimuli for different fixed-viewing conditions.
Observers had to detect to which CIFAR-10 category did the image belong to. 
The categories were linked to key presses of their most important letters \emph{(A)irplane, a(U)tomobile, (B)ird, (C)at, d(E)er, (D)og, (F)rog, (H)orse, (S)hip} and \emph{(T)ruck}. The stimulus which had CIFAR-10 grayscale images were of size 32x32 pixels and were scaled to 190x190 pixels for optimal viewing based on the fact that lower dimension images are better viewed from a distance or if their dimensions are smaller ~\cite{Pelli_1999}.

\begin{figure}[h!]
\begin{center}
\centerline{\includegraphics[width=\columnwidth]{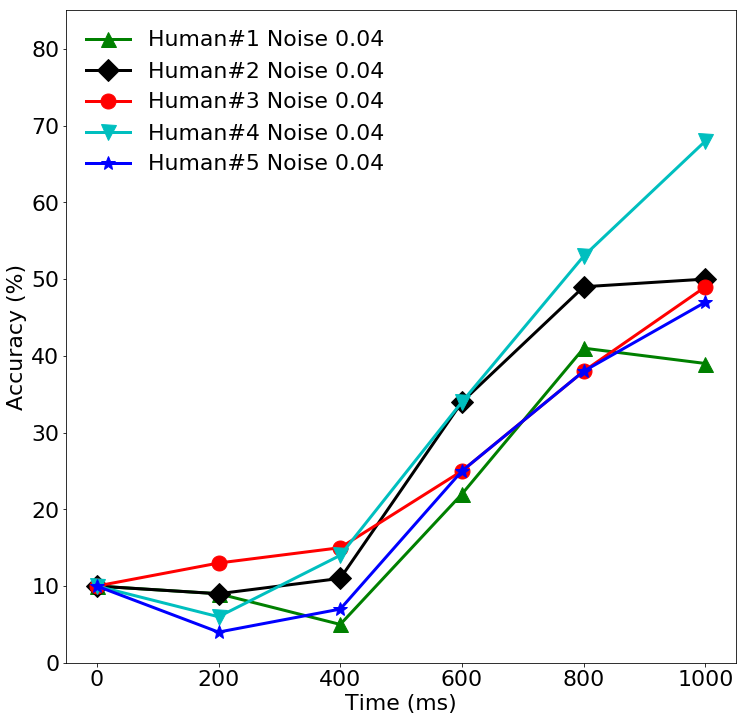}}
\caption{{\fontfamily{phv}\fontsize{9}{10}\selectfont
Human accuracy as a function of required response time. Noise with SD 0.04 was added to each image. Accuracy tends to increases with response time. Accuracy at 0 ms was not measured and is assumed to be at chance (10\%)}} 
\label{fig:humans}
\end{center}
\end{figure}

\paragraph{Design.}
Since 150 ms is the visual processing time needed to understand a stimulus ~\cite{Thorpe_1996}, the survey was designed on 5 fixed viewing conditions of 200 ms, 400 ms, 600 ms, 800 ms, and 1000 ms with a tolerance of $\pm 100$ ms for discarding invalid trials. Each fixed viewing condition had 100 trials with decreasing time from 1000 ms to 200 ms. At the end of each time-limit, the observer had to press at the beep of 60 ms to enter their category via key-press. At the end of the stimuli, the observer was given feedback by mentioning if they were quick, slow or perfect while pressing the key. Thus for each observer, there were 5 blocks of 100 trials each. 
\paragraph{Procedure.}
Observers were asked place their hands on the keyboard while being aware of the ten identifiers (A: Airplane, C: Cat etc). All of the observers placed their index fingers on F and J to remember the mapping of the keys. Observers were instructed to answer at the beep as frequently as possible because time and accuracy were both recorded. Feedback appeared at the centre of the screen to make them to get them used to the timing of the beep after each trial, and pressing spacebar showed the next stimulus. The 1000 CIFAR-10 grayscale images were randomly sampled from 10,000 test images while designing the survey and each participant was shown 100 randomly sampled images during each block of fixed viewing condition. The stimulus would disappear after time limit for each condition and each participant had to learn to answer at the beep. 
\paragraph{Human performance.}
Figure~\ref{fig:humans} plots human accuracy on CIFAR-10 as a function of reaction time. At 1000 ms, most observers had accuracy about 40\% to 50\%, except for Human\#4, who was more accurate. 

\paragraph{Model architecture.}
The architecture of MSDNet is illustrated in Figure~\ref{fig:layout}. 
The first layer includes vertical connections on $S$ scales. The feature maps at coarser scales are obtained via convolutions and down sampling which generates the first representations for the network to train on. Coarse features are important for getting better performance in intermediate classifiers.
Subsequent layers process a concatenation of transformed feature maps from previous scales $s$ and $s-1$ (if $s > 1$). Classifier layers in MSDNet use a dense connectivity pattern, where all features $\left[x_1^{S},\ldots,x_\ell^{S}\right]$ are used at layer $\ell$.  Dense connectivity~\cite{Huang_2017_CVPR} is determined to be important to maintain a high accuracy of the final classifier and early exit classifiers and is a crucial feature of the network~\cite{huang2017multiscale}. A cascade of intermediate classifiers is used to benefit from this dense connectivity as it allows layers to bypass features optimized for the short-term. MSDNet conserves feature representation at multiple scales, and all the classifiers use coarse-level features. This is done to reduce high error rates in intermediate and final classifiers. 
MSDNet uses a cross entropy loss function across all classifiers $L(f_k)$. During training, a weighted cumulative loss is used to optimize the parameters of the network. This loss is defined by:  $1/\mathcal{|D|}\sum\nolimits_{(\mathbf{x},y)\in\mathcal{D}} \sum\nolimits_{k}w_k L(f_k)$, where $\mathcal{D}$ is the training set and $w_k\!\geq\! 0$ is the weight of $k$-th classifier. To use the anytime setting, a test image is passed through the network and a prediction is expected when the FLOP budget is exhausted. The budget $C_k$ to obtain the prediction at each classifier $k$ is fixed and determined by the network architecture and input image size. This trains the network end-to-end to optimize the loss function based on all the classifiers of the network. Early-exits only take place during inference. For our experiments, we use a 15-layer deep MSDNet network with 7 classifiers and evaluated the top-1 accuracy during anytime setting. One important difference from the original MSDNet is the way bottleneck layer implemented. We use a 1-1-1 setting as compared to 1-2-4 setting in the original MSDNet. This is to add constraints to the original network and inhibit the ability of the initial layers to reach higher accuracy. The first early classifier is placed after 3 layers and rest of classifiers are placed after every 2 layers. The first block contains scales of 8, 14 and 16 which sets up representations for the layers in the next blocks. The budgets $C_k$ vary from $3.56\times10^{6}$ to $12.21\times10^{6}$ FLOPs. The main difference of this model as compared to MSDNet is reducing the bottleneck factor and number of scales utilized in the initial layers of the network.  

\paragraph{Training details.}
 MSDNet is configured to use 7 blocks for getting top-1 accuracy at the end of each block. We convert the CIFAR-10 dataset into grayscale images. Data augmentation based on standard techniques mentioned in ~\cite{huang2017multiscale} is applied: during training, images are horizontally flipped with probability 0.5, normalization based on channel means and standard deviation is also done. Noise in perception experiments is used for assessing unpredicable variation in some aspect of stimulus~\cite{pelli2015}, and we attempt to model the same effect in our experiments. We apply grayscale Gaussian noise~\footnote{Grayscale Gaussian noise indicates that noise samples generated are same across each channel.} with 0.0 mean and standard deviation from 0.0 to 0.15 with a step of 0.02 with equal probability to image batches during training. The model is trained for 300 epochs with stochastic gradient descent optimizer and a mini-batch batch size $64$. All results hereafter are reported with these training settings.

\paragraph{Inference.}
MSDNet uses confidence scores to compare each test example with an exit probability and trade-off FLOPs for easy and hard examples. If an image gets assigned high confidence score early on, the model exits with a smaller $k$, utilizing less FLOPs. During inference, Gaussian noise with 0.0 mean and different standard deviation values is generated and applied to test images to model difficulty. Standard deviation is used to generate fine-grained variation in test accuracy. 

\section{Results and Discussion}
\begin{figure}[t] 
\begin{center}
\centerline{\includegraphics[width=\columnwidth]{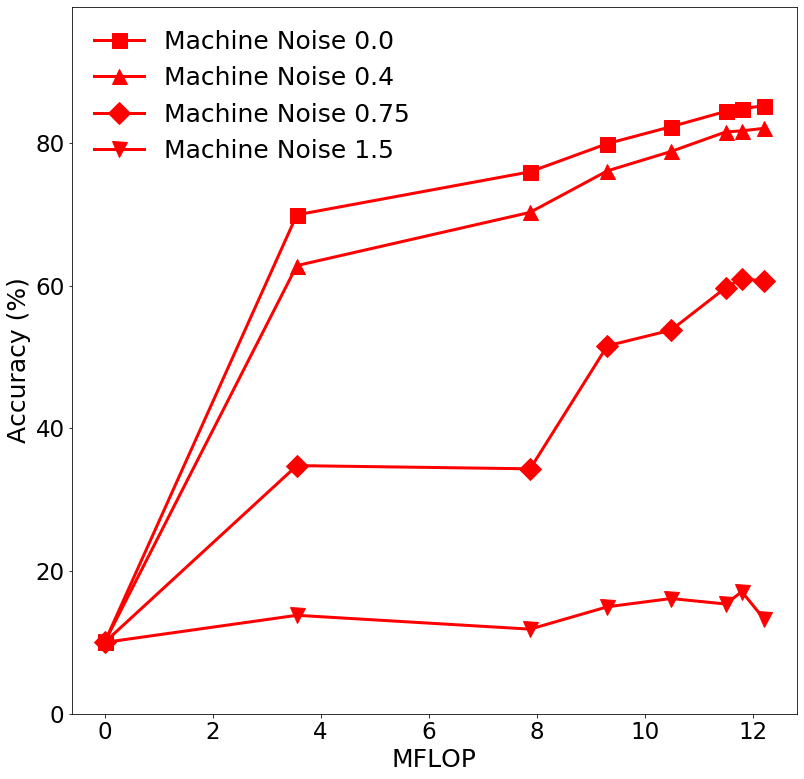}}
\caption{{\fontfamily{phv}\fontsize{9}{10}\selectfont
Accuracy of anytime prediction by our network on CIFAR-10 test images with several amounts of added noise, specified by SD. Each point is the accuracy of a particular exit against its computational cost (MFLOP). The noise is white, zero mean, gaussian with specified standard deviation. Accuracy at 0 MFLOP was not measured and is assumed to be at chance (10\%). We estimate an equivalent input noise of 0.6 SD. Adding noise less than the equivalent input noise has little effect. }} 
\label{fig:analysis1}
\end{center}
\vskip -0.2in
\end{figure}

We now present a summary of our experimental results.
\paragraph{Testing MSDNet.} Figure~\ref{fig:analysis1} plots measured accuracy at each exit (which determines MFLOPs) at each noise level.  The noise is added to an image whose pixels are in the range 0.0 to 1.0. Further computation (more MFLOP), from exit to exit, improves accuracy at every noise level, and increasing noise reduces accuracy at every exit (MFLOPs). 
Note that the network does make errors even without any noise, and the similar shape of curves with and without noise suggests that, following the advice of ~\cite{pelli1999use}, one might model the effects of noise by imputing an intrinsic noise within the network, equivalent to an added noise of perhaps 0.5 sd. Adding noise that is weaker than 0.5 has very little effect.

\begin{figure}[t!]
\begin{center}
\centerline{\includegraphics[width=\columnwidth]{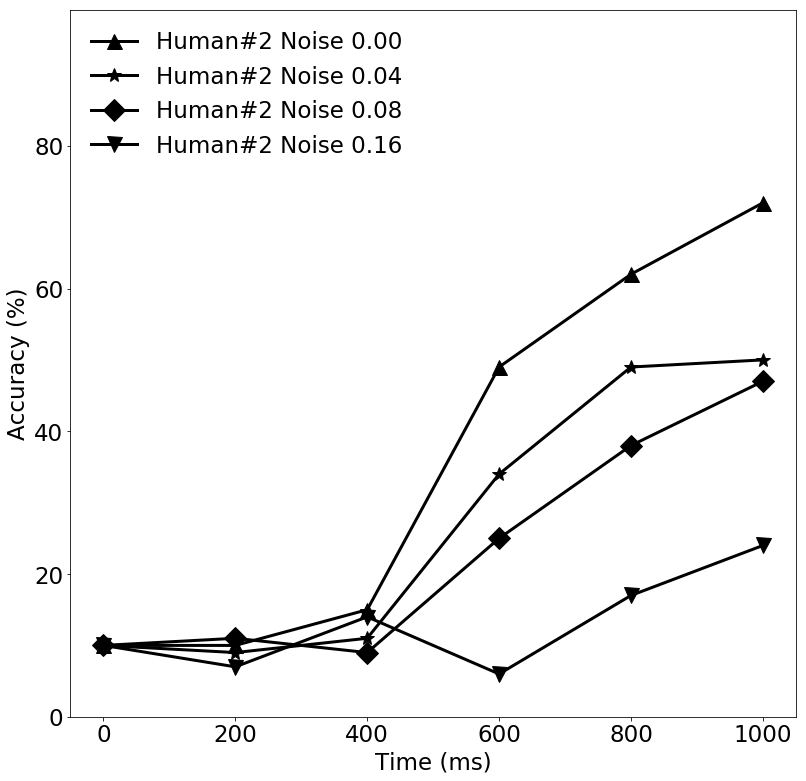}}
\caption{{\fontfamily{phv}\fontsize{9}{10}\selectfont
Human\#2 accuracy at noise levels of 0.0, 0.04, 0.08, and 0.16. Accuracy tends to fall as noise is increased. For times up to 400 ms the accuracy is close to chance. Accuracy at 0 ms was not measured and is assumed to be at chance (10\%). We estimate an equivalent input noise of 0.04 SD. Adding noise less than the equivalent input noise has little effect.}} 
\label{fig:humans1}
\end{center}
\end{figure}

\paragraph{Testing humans.}
We followed the same procedure for human experiments. Fig.~\ref{fig:humans1} shows performance of Human\#2 for images at several values of noise SD and many response times. Each block tested the observer with 500 images, training the observer to respond at the specified duration.
Fig.~\ref{fig:humans1} shows the effect of noise on human\#2. Accuracy tended to increase with response time, and tended to fall with increasing noise SD. The human too makes mistakes without noise and we estimate an equivalent input noise of perhaps 0.04.

\begin{figure}[h!] 
\begin{center}
\centerline{\includegraphics[width=\columnwidth]{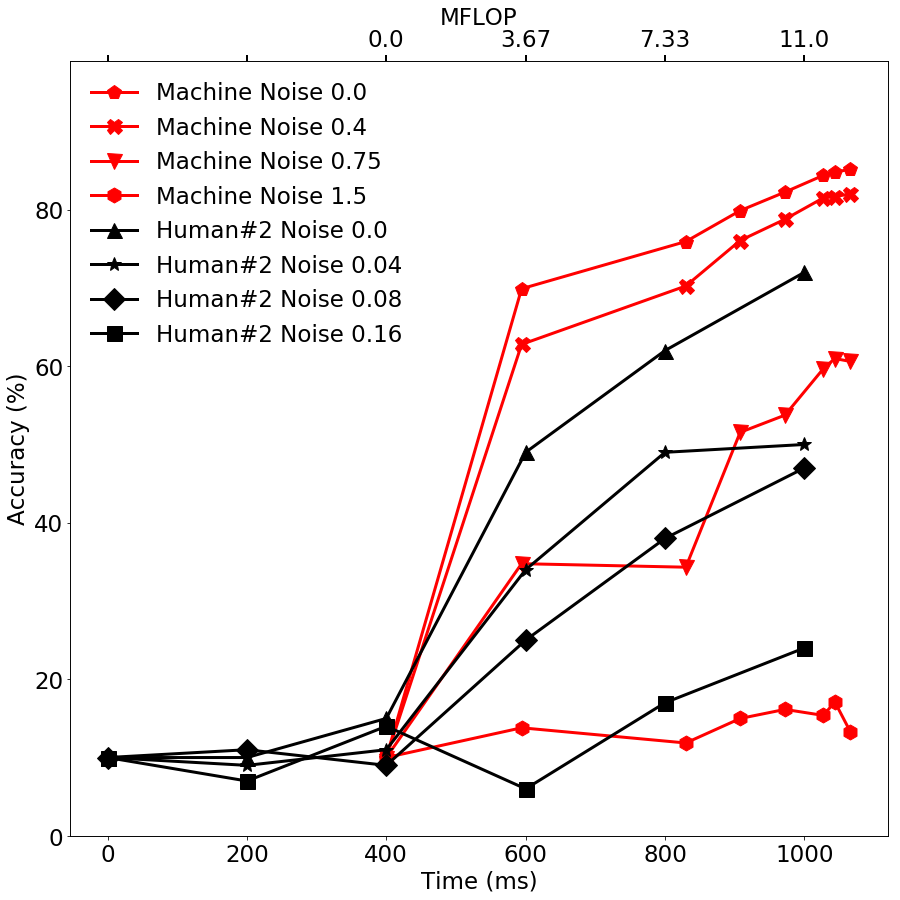}}
\caption{{\fontfamily{phv}\fontsize{9}{10}\selectfont
Comparison of human\#2 and MSDNet. Accuracy is plotted vs. ms (at bottom) or MFLOP (at top). Time in ms and MFLOPs are related by Equation ~\ref{eq:fit}. Small amounts of noise (less than the equivalent input noise) have little effect. The second strongest noise SD for the network was 0.75, and its curve seems to correspond to an intermediate noise SD of 0.06 for the human.  The ratio, 0.06/0.75=0.08 indicates that human efficiency (ratio of noise variances for equal performance with equal signals) is $0.08^2=0.6$\% that of MSDNET. 
}}
\label{fig:comparison}
\end{center}
\end{figure}

\paragraph{Comparing humans and machines.}
To compare human and machine performance we need to suppose a correspondence between MFLOPs and time (ms). 
Figure~\ref{fig:comparison} plots human accuracy vs time and network accuracy vs FLOPs.
Accuracy is at chance 10\% until processing begins at zero time or zero FLOP. The MFLOP scale \textit{F} (on the top of the graph) is linearly related to time scale T (in ms) by Equation ~\ref{eq:fit}. 
\begin{equation}
 F = (11/600) (T-400)
\label{eq:fit}   
\end{equation}Both the scale factor 11/600 (between MFLOPs and ms) and the 400 ms offset (to account for delays in retinal processing and motor planning and execution) were adjusted by eye to align the the human and machine results.

The network and human curves are collected at different noise levels, but are otherwise quite similar. The need for much higher noise to bring the network accuracy down to human performance indicates that the network is much more efficient. \citeauthor{tjan1995human} reported efficiencies of 3\% to 6\% for human recognition of one of four grayscale images of 3D objects in noise. Relative \textit{Efficiency} is the ratio of noise variances that allows the test observer  to achieve the same performance with the same signal energy as the reference observer. Thus, Figure~\ref{fig:comparison} indicates a relative efficiency of roughly 0.6\%. Both man and machine were little affected by the weakest noise applied, presumably because it was small relative to the  equivalent input noise of that observer (man or machine).
\section{Conclusion and Future Work}
Comparing human and MSDNet accuracy in classifying CIFAR-10 images in added Gaussian noise, we find that the network equivalent input noise SD is 15 times higher than human, and that human efficiency is only 0.6\% that of the network. When appropriate amounts of noise are present to bring the two observers (human and network) into the same accuracy range, they show very similar dependence on duration or FLOPS, i.e. very similar speed-accuracy tradeoff. We conclude that Anytime classification (i.e. early exits) is a promising model for human reaction time in recognition tasks.

In future work, we will apply this network to modeling word reaction times of dyslexic participants.

\section{Acknowledgments}
We thank Augustin Burchell for putting together our remote testing stack using Lab.js \cite{henninger_felix_2020_3953072} and JATOS \cite{Lange_2015}. Thanks to Lisa Levinson for advice on human reaction time. We also thank Kuan-Lin Liu, Sam Shen, and Augustin Burchell for participating as observers. We gratefully acknowledge support from the Moore Sloan Foundation, NYU Center for Data Science, and NIH grant R01 EY027964 to DGP. 

\bibliography{anytime_pred}
\bibliographystyle{aaai21.bst}

\end{document}